\documentclass[10pt,twocolumn,letterpaper]{article}

\usepackage{multirow}
\usepackage{cvpr}              

\usepackage[dvipsnames]{xcolor}

\usepackage{hyperref} 

\title{\emph{MuTri}: Multi-view Tri-alignment for OCT to OCTA 3D Image Translation}
\author{
	{Zhuangzhuang Chen}$^{1}$
	~~~ {Hualiang Wang}$^{1}$
	~~~ {Chubin Ou}$^{2}$
	~~~ {Xiaomeng Li}$^{1}$\thanks{Corresponding author.} \\
	\textsuperscript{1}  Department of Electronic and Computer Engineering, The Hong Kong University of Science and Technology\\ 
        \textsuperscript{2} Department of Radiology, Guangdong Provincial People's Hospital, Southern Medical University \\ 
        \texttt{\small{eezzchen@ust.hk, hwangfd@ust.hk, cou@connect.ust.hk, eexmli@ust.hk
}}}

\begin{document}
\maketitle     
\begin{abstract}
Optical coherence tomography angiography (OCTA) shows its great importance in imaging microvascular networks by providing accurate 3D imaging of blood vessels, but it relies upon specialized sensors and expensive devices. For this reason, previous works show the potential to translate the readily available 3D Optical Coherence Tomography (OCT) images into 3D OCTA images.  However, existing OCTA translation methods directly learn the mapping from the OCT domain to the OCTA domain in continuous and infinite space with guidance from only a single view, i.e., the OCTA project map, resulting in suboptimal results. To this end, we propose the multi-view Tri-alignment framework for OCT to OCTA 3D image translation in discrete and finite space, named \emph{MuTri}. In the first stage, we pre-train two vector-quantized variational auto-encoder (VQVAE) by reconstructing 3D OCT and 3D OCTA data, providing semantic prior for subsequent multi-view guidances. In the second stage, our multi-view tri-alignment facilitates another VQVAE model to learn the mapping from the OCT domain to the OCTA domain in discrete and finite space. Specifically, a contrastive-inspired semantic alignment is proposed to maximize the mutual information with the pre-trained models from OCT and OCTA views, to facilitate codebook learning. Meanwhile, a vessel structure alignment is proposed to minimize the structure discrepancy with the pre-trained models from the OCTA project map view, benefiting from learning the detailed vessel structure information. We also collect the first large-scale dataset, namely, OCTA2024, which contains a pair of OCT and OCTA volumes from 846 subjects. \hypertarget{label}{\href{https://github.com/xmed-lab/MuTri}{Our codes and datasets are available at: https://github.com/xmed-lab/MuTri}}.
\end{abstract}
    
\section{Introduction}
\label{sec:intro}

\begin{figure}[t]
\centering    
\includegraphics[scale=0.83]{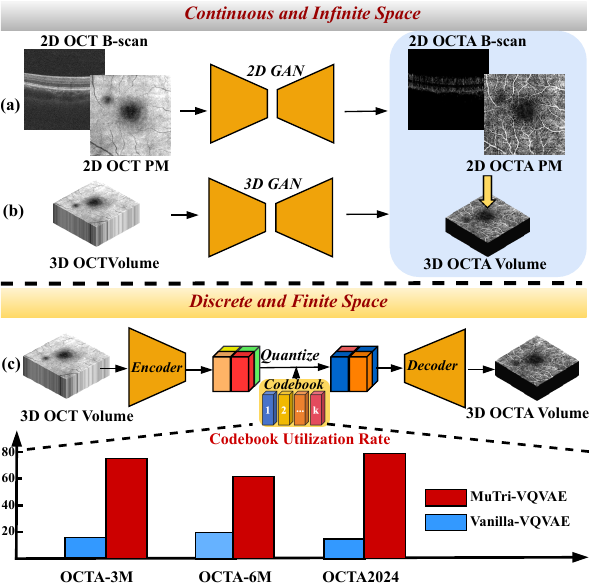}
\caption{(a)~The existing OCTA translation methods learn the mapping in the infinite and continuous space: (a) 2D OCT projection map (PM) or B-scan images as input, (b) 3D OCT volume as input with only single view guidance from  2D OCT PM. (c) A vanilla VQVAE involves a lower codebook utilization rate for OCT to OCTA 3D image translation tasks, resulting in a less informative codebook. Note that, the codebook utilization rate is computed by using training samples from three datasets.}
\label{fig:mo}
\end{figure}


Optical Coherence Tomography (OCT) acts as a non-invasive imaging technology for providing highly detailed in-depth retinal pathologies  \cite{badhon2024quantitative}. Thus, it has been widely adopted in ophthalmic clinical care. However, as a non-dynamic solution, OCT cannot visualize blood flow information such as blood vessel caliber or density and remains limited to capturing structural information \cite{fujimoto2016development}. For this cause, OCT angiography (OCTA) is developed to produce volumetric data from choroidal and retinal layers and fill the above information gap \cite{chalam2016optical,kim2022integrated}. OCTA has shown its great potential in detecting diabetic retinopathy \cite{alam2020quantitative}, age-related macular degeneration \cite{burlina2017automated}, Glaucoma \cite{hollo2016vessel} and several other retinal diseases \cite{alam2017quantitative,hsieh2019oct}. Despite these advantages, OCTA data involves high costs including additional hardware and software modifications when compared with OCT data \cite{lin2024deep}. Meanwhile, OCTA devices have not been well-popularized than OCT devices. Considering this, a promising solution to this problem can be the utilization of deep learning methods to produce OCTA modality from the widely-used and cost-effective OCT modality \cite{li2024vessel,lin2024deep,badhon2024quantitative}, to achieve low-cost upgrading of OCT devices.

Existing OCT to OCTA  translation approaches \cite{lee2019generating,li2020deep,lin2024deep,li2024vessel} first extract the features of the OCT domain into a continuous, infinite space, and subsequently expand these features into the OCTA domain. Previous works \cite{lee2019generating,li2020deep,pan2022multigan} mainly focus on translating 2D OCT projection map (PM)/B-scan images into their corresponding 2D OCTA projection map/B-scan images by using 2D Generative Adversarial Network (GAN), as shown in Fig~\ref{fig:mo} (a). Notably, these methods only focus on 2D input, i.e.,  either B-scan or projection map, failing to capture depth information in the 3D volume. To overcome this, as shown in Fig.~\ref{fig:mo} (b), a recent method, TransPro \cite{li2024vessel} takes the 3D OCT volume as input, and then generates 3D OCTA volume via a 3D GAN with single view guidance from a pre-trained 2D projection map translation model and a vascular segmentation model. However, this approach has several major limitations. First, it is challenging to learn the direct mapping from the OCT to the OCTA domain in an infinite and continuous space, resulting in limited translation quality of OCTA. Second, this method depends on a pre-trained vascular segmentation model for network guidance, which is difficult to acquire due to the substantial effort required to develop well-annotated vascular segmentation datasets.




To overcome these limitations, a promising approach is to utilize Vector Quantization (VQ) in the translation process. This method can significantly reduce mapping uncertainty by learning a codebook within a discrete and finite space, as shown in the general 2D image domain~\cite{razavi2019generating,esser2021taming}.
However, simply using a vanilla VQ Variational Autoencoder (VQVAE) often leads to a less informative codebook with lower utilization, as shown in Figure~\ref{fig:mo}(c). This ultimately results in unsatisfactory results, as seen in the ``Vanilla VQVAE'' results in Table~\ref{tab:abla}.
Its limitation can be summarized as two-fold: (i) Due to the large domain gap between the OCT and OCTA domains, it is challenging to learn informative unquantized OCT features and quantized OCTA features, making it difficult to develop a well-optimized codebook that can effectively transform unquantized OCT features into well-quantized OCTA features. (ii) The vanilla VQVAE does not effectively utilize the pre-trained OCT and OCTA codebooks for the downstream transformation from OCT to OCTA.

To this end, we propose a novel \emph{MuTri}, which is a two-stage OCT to OCTA 3D image translation framework including the pertaining of OCT and OCTA reconstruction models in the first stage, and the second stage leverage the pre-trained models to guide the fully OCTA translation process from three views: 3D OCT, 3D OCTA, and 2D OCTA project maps. Our key idea is that the pre-trained OCT and OCTA reconstruction models can provide high-quality domain features, thus providing multi-view guidance that allows the translation model to learn informative OCT features, OCTA features, and codebooks. Specifically, the contrastive-inspired semantic alignment is proposed to maximize the mutual information with the pre-trained models from both 3D OCT and 3D OCTA views, to encourage the codebook to explore more new codewords. Meanwhile, inspired by the fact that the vessel structure information can be highlighted in the 2D OCTA project map, vessel structure alignment considers patch-level semantic correlation and minimizes the structure discrepancy with the pre-trained OCTA model from the 2D OCTA project map view. By enforcing this proximity, the second stage will benefit from learning the detailed vessel structure information.

The contributions can be summarized as four-fold:
\begin{itemize}

\item[$\bullet$] To the best of our knowledge, this is the first work to learn the OCT to OCTA mapping in discrete and finite space with multi-view guidances: 3D OCT,  3D OCTA, and 2D OCTA projection map. To this end, we propose the multi-view tri-alignment framework to enable the optimization of codebook learning for OCTA translation tasks.

\item[$\bullet$] Contrastive-inspired semantic alignment is proposed to maximize the mutual information with the pre-trained models from 3D OCT and 3D OCTA views, to encourage the codebook to explore more new codewords. 

\item[$\bullet$] Vessel structure alignment is proposed to achieve vessel structure consistency by considering patch-level semantic correlation from the 2D OCTA projection map view. 

\item[$\bullet$] We propose the first large-scale OCT to OCTA 3D translation dataset, named OCTA2024, which is collected from real-world 848 subjects, to promote related research by serving as a new OCTA translation benchmark. Extensive experiments illustrate that our \emph{MuTri} achieves state-of-the-art performance quantitatively and qualitatively.

\end{itemize}

\section{Related work}
\label{sec:formatting}

\subsection{OCT to OCTA Image Translation}
Automatic OCT to OCTA image translation methods act as a promising solution for the acquisition of challenging OCTA modality. It aims to translate an easily obtainable OCT image modality to a more challenging OCTA modality \cite{mcnaughton2023machine}. According to the existing research \cite{li2024vessel}, we divide the current OCT to OCTA translation methods into two groups according to the types of input images. The first group methods focus on generating 2D OCTA B-scan images from paired 2D OCT 
B-scan images \cite{lee2019generating,li2020deep,zhang2021texture} or 2D OCTA projection maps from paired 2D OCT projection maps \cite{pan2022multigan,badhon2024quantitative}. More specifically, Lee et al. \cite{lee2019generating} develop an encoder-decoder model to achieve 2D OCT to OCTA B-scan image translation. Later, Zhang et al. \cite{zhang2021texture} consider the texture features in OCT B-scans and propose Texture-UNet for improving the above translation process. Inspired by the GAN models, Li et al.~\cite{li2020deep} propose AdjacentGAN that can effectively prompt the translation of the OCTA B-scan image by leveraging limited contextual information from neighboring OCT slices. Unlike these works, MultiGAN \cite{pan2022multigan} explores the potential of the translation of OCT projection maps to OCTA projection maps. Then, Badhon et al.~\cite{badhon2024quantitative} further implement a generative adversarial network framework that includes a 2D vascular segmentation model and a 2D OCTA image translation model. The second group methods aim to leverage the abundant information in 3D volume and achieve 3D OCT to OCTA translation. To achieve this, TransPro \cite{li2024vessel} introduces vascular segmentation as an auxiliary task to enhance the quality of vascular regions in the translated OCTA images. However, this method heavily relies on the pre-trained vascular segmentation model, which is hard to access due to the huge pixel-level labeling costs.

 In contrast to these methods that learn the OCT to OCTA mapping in the infinite and continuous space, herein, this work advances Vector-Quantization (VQ) into OCTA translation tasks and aims to learn a codebook in discrete and finite space. Meanwhile, this work focuses on multi-view tri-alignment including 3D OCT, 3D OCTA, and 2D OCTA projection map, and does not rely on any additional costly vascular annotations.

\subsection{General Image to Image Translation}
Considering the fact that our\emph{MuTri} serves as an image-to-image translator, we would like to give a brief review of related research on this scope. These methods aim to translate an input image from a given domain to another domain \cite{xia2024diffusion}. Given the paired training samples, Pix2Pix \cite{isola2017image} leverages the conditional generative adversarial network (CGAN) to inject the information of the input domain and transfer it into the output domain. Meanwhile, Tsit \cite{jiang2020tsit} designs a coarse-to-fine fashion for feature transformations. However, these methods still suffer from training instability and the severe mode collapse issue \cite{li2020diversity,li2024scenedreamer360}. Notably, VQ-I2I \cite{chen2022eccv} propose a vector quantization technique into the image-to-image translation framework with the disentangled style representation. Recently, several studies have shown the potential of the Diffusion probabilistic model (DPM) for image-to-image translation masks. For example, Palette \cite{saharia2022palette} introduces a DPM framework that aims to inject the input into each sampling step for refinement. PITI \cite{wang2022pretraining} and DiffIR \cite{xia2023diffir} adapts a pre-trained DPM model to accommodate various kinds of image-to-image translation. Despite many techniques that have been developed to significantly reduce the number of denoising steps during diffusion, their time-consuming generation process has not been well-addressed \cite{xia2024diffusion}. 

Unlike the above methods, we focus on facilitating codebook learning in discrete and finite space by aligning with the pre-trained models that can provide multi-view priors.

\subsection{Contrastive Learning for Image Translation}
Since this paper is amount to leverage contrastive learning for image generation, we give a brief review of the related works. Generally speaking, contrastive learning acts as a well solution for learning better feature representations and can be applied on various tasks. For example, He et al. \cite{he2020momentum} propose MoCo to achieve unsupervised visual representation learning. Later,  Tian et al. \cite{tian2020contrastive} greatly extended MoCo by considering two more input views. These methods can be further connected to the Noise Contrastive Estimation \cite{gutmann2010noise}. Then, the encoder can learn image-level representation rather than pixel-level generation. To achieve representation learning for pixel-level generation, CUT \cite{park2020contrastive} is the first work that utilizes patch-wise contrastive estimation on unpaired translation
tasks for image to image translation tasks. Jeong et al.~\cite{jeong2021memory} enhance the discrimination ability of memory with the proposed feature contrastive loss. However, it involves huge computation burdens as the memory bank gets outdated quickly in a few passes. Later,  MCL \cite{gou2023multi} designed multi-feature contrastive learning to construct a patch-wise contrastive loss using the feature information of the discriminator output layer. Recently, Zhao et al.~\cite{zhao2024spectral} propose the dual contrastive regularization for image translation framework.

We further carry forward the idea of contrastive learning for codebook learning under the OCT to OCTA translation task. Specifically, we maximize the mutual information between the translation model (OCT to OCTA) and pre-trained reconstruction models (OCT and OCTA), promoting the codebook to explore more new codewords, instead of simply learning better features as in the existing works.

\begin{figure*}[t]
\centering    
\includegraphics[scale=0.5]{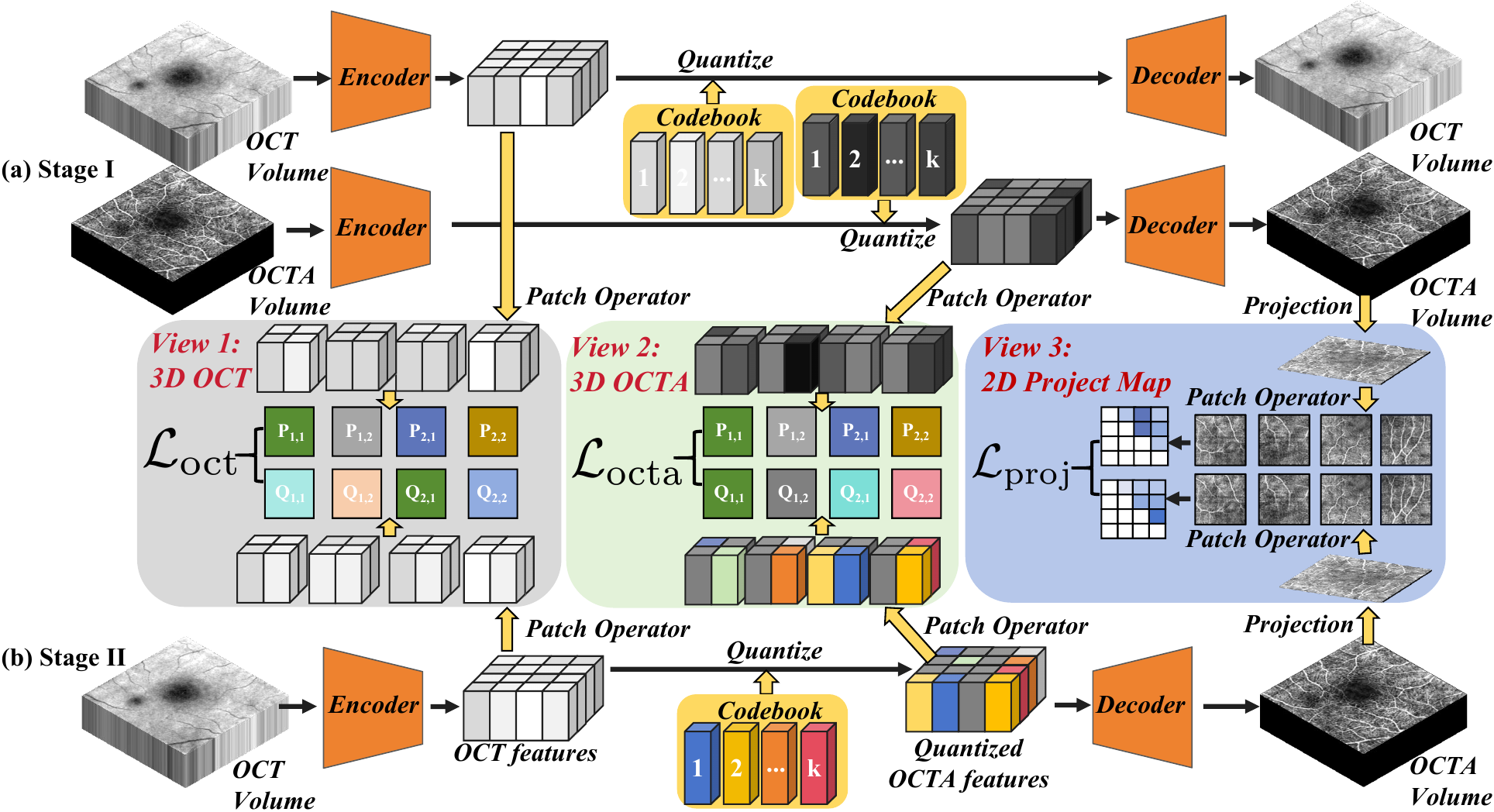}
\caption{The overall pipeline of \emph{MuTri}. It consists of two stages to facilitate OCT to OCTA 3D  image translation. (a) Stage 1 employs two VQVAE pre-trained on the OCT and OCTA volumes, to provide multi-view guidances: 3D OCT, 3D OCTA, and 2D OCTA projection map. (b) Stage 2 utilizes another VQVAE that takes OCT volumes as input, to reconstruct the OCTA volumes under our contrastive-inspired semantic alignment from 3D OCT and OCTA views and vessel structure alignment from 2D OCTA projection map view.}
\label{fig:overview}
\end{figure*}
\section{Method}

\subsection{Framework Overview} 
In this paper, we propose a novel framework, called Multi-view Tri-alignment (\emph{MuTri}) for OCT to OCTA 3D image translation. As illustrated in Fig.~\ref{fig:overview}, the training of \emph{MuTri} consists of two stages. In the first training stage (see Fig.~\ref{fig:overview}~(a)), we pre-train two VQVAEs by individually reconstructing the OCT and OCTA volumes which provide multi-view guidance for the subsequent stages. Then, in the second training stage (see Fig.~\ref{fig:overview}~(b)), we train another VQVAE with the help of multi-view guidances: 3D OCT,  3D OCTA, and 2D OCTA projection map, to enhance codebook learning in OCT to OCTA translation tasks. The trained VQVAE can easily translate OCT into OCTA volumes in the inference phase. In the following section, we will discuss more details of \emph{MuTri}.

\subsection{Stage I: Multi-modality Pre-training}
\label{sec:stageI}
Given the 3D OCT volume $X_{\text{oct}}$ and OCTA volume $X_{\text{octa}}$, two 3D convolutional encoder $\text{E}_{\text{oct}}$ and $\text{E}_{\text{octa}}$ maps the input $X_{\text{oct}}$ and $X_{\text{octa}}$ to the unquantized feature vector $\mathbf{u}_{\text{oct}}$ and $\mathbf{u}_{\text{octa}}$ as following :
\begin{equation}
\mathbf{u}_{\text{oct}} ={\text{E}}_{\text{oct}}(X_{\text{oct}}),~~~~\mathbf{u}_{\text{octa}} =\text{E}_{\text{octa}}(X_{\text{octa}}).
\end{equation}
Next, we built two codebooks $\mathcal{Z}_{\text{oct}},\mathcal{Z}_{\text{octa}} \in \mathbb{R}^{N \times d}$ to capture multi-modality information of OCT and OCTA dataset, where $N$ is the number of entries in the codebook. And, $d$ is the dimension of each entry. Then, we can discretize the distribution of unquantized feature vector to get the quantized features $\mathbf{q}^{\mathcal{T}}_{\text{oct}}$, $\mathbf{q}^{\mathcal{T}}_{\text{octa}}$ by the following vector quantization operator $\text{VQ}_{\mathcal{Z}}(\cdot)$:
\begin{equation}
\text{VQ}_{\mathcal{Z}}({z}):= \operatorname{arg min}_{\mathbf{z}_{k} \in \mathcal{Z}} \left\|{z}-\mathbf{z}_{k} \right\|,
\label{eq:vq}
\end{equation}   
where $\mathbf{z}_{k}$ is the $k$-th entry in the codebook $\mathcal{Z}$. Then, each vector in $\mathbf{u}_{\text{oct}}$ and $\mathbf{u}_{\text{octa}}$ can be replaced with its nearest neighbor entry in the codebook $\mathcal{Z}_{\text{oct}}$ and $\mathcal{Z}_{\text{octa}}$ via Eq.~\ref{eq:vq}, resulting in the quantized feature vector $\mathbf{q}_{\text{oct}}$ and $\mathbf{q}_{\text{octa}}$: 
\begin{equation}
\mathbf{q}_{\text{oct}} = \text{VQ}_{\mathcal{Z}_{\text{oct}}}(\mathbf{u}_{\text{oct}}),~~\mathbf{q}_{\text{octa}} = \text{VQ}_{\mathcal{Z}_{\text{octa}}}(\mathbf{u}_{\text{octa}}).   
\end{equation}
The decoder $\text{D}_{\text{oct}}$ and  $\text{D}_{\text{octa}}$ reconstructs the quantized vector $\mathbf{q}_{\text{oct}}$ and $\mathbf{q}_{\text{octa}}$ back into the volume $\hat{X}_{\text{oct}}$ and $\hat{X}_{\text{octa}}$ as: 
\begin{equation}
\hat{X}_{\text{oct}}= \text{D}_{\text{oct}}(\mathbf{q}_{\text{oct}}),~~~\hat{X}_{\text{octa}} =  \text{D}_{\text{octa}}( \mathbf{q}_{\text{octa}} ).
\label{eq:recon}
\end{equation}
Finally, codebook $\mathcal{Z}_{\text{oct}}$, encoder $\text{E}_{\text{oct}}$, and decoder $\text{D}_{\text{oct}}$ are jointly optimized with the following loss function:
\begin{equation}
\begin{aligned}
\mathcal{L}_{\text{VQVAE}}(\mathcal{Z}_{\text{oct}}, \text{E} _{\text{oct}},  & \text{D} _{\text{oct}}) =  \left\|  X_{\text{oct}} - \hat{X} _{\text{oct}}  \right\| +  \\
&  \left\|   \mathbf{sg}[ \mathbf{u} _{\text{oct}} ] - \mathbf{q} _{\text{oct}}    \right\| +  \left\|   \mathbf{sg}[\mathbf{q} _{\text{oct}} ] - \mathbf{u} _{\text{oct}}  \right\|,
\end{aligned}
\label{eq:vqvae}
\end{equation}    
where $\mathbf{sg}[\cdot]$ indicates the stop gradient operator. The first item in Eq.~\ref{eq:vqvae} optimizes the encoder and decoder to enforce the reconstructed volume close to the original OCT volume. The second item in Eq.~\ref{eq:vqvae} provides gradients to the codebook $\mathcal{Z}_{\text{oct}}$. To incentivize the encoder $\text{E} _{\text{oct}}$ to commit to the codebook, a third term is added to update the encoder parameters. In other words, it enforces the latent feature $\mathbf{u} _{\text{oct}} $ to be close to the nearest neighbor entry in  $\mathcal{Z}_{\text{oct}}$. Meanwhile, $\mathcal{Z}_{\text{octa}}$, $\text{E} _{\text{octa}}$, and $\text{D} _{\text{octa}}$ are also optimized by the loss function $\mathcal{L}_{\text{VQVAE}}(\mathcal{Z}_{\text{octa}}, \text{E} _{\text{octa}}, \text{D} _{\text{octa}})$ as formulated in Eq.~\ref{eq:vqvae}.

After training Stage-I VQVAE models,  we are allowed to obtain unquantized features from OCT volumes and quantized features from OCTA volumes. In the following Sec~\ref{sec:stageII}, we discuss how to leverage those informative unquantized/quantized prior features to promote OCT to OCTA 3D image translation tasks. 

\subsection{Stage II: Multi-view Tri-alignment}
\label{sec:stageII}
The success of OCT to OCTA image translation via a vector quantization approach lies in a well-learned codebook that allows us to quantize the continuous OCT features into discrete OCTA representations. However, due to the huge domain gap between OCT and OCTA modality, training a VQVAE alone often leads to model collapse, where the model gets stuck in local optima and fails to optimize the codebook, resulting in a lower codebook utilization rate, see Fig.~\ref{fig:mo}~(c). Considering this, we \textbf{first} propose contrastive-inspired semantic alignment to maximize the mutual information with the pre-trained models from 3D OCT and 3D OCTA views, to encourage the codebook to explore more new codewords. Motivated by the fact that vessel structure information can be highlighted in the 2D OCTA project map, we \textbf{then} propose vessel structure alignment to achieve vessel structure consistency by considering patch-level semantic correlation from the 2D OCTA projection map view. 

\noindent\textbf{Contrastive-inspired semantic alignment:} Given a translation VQVAE model that contains encoder $\text{E}_{\text{oct2octa}}$, codebook $\mathcal{Z}_{\text{oct2octa}}$, and decoder $\text{D}_{\text{oct2octa}}$, it takes the 3D OCT volume as input, and then generates 3D OCTA volume. Specifically, we first input $X_{\text{oct}}$ to encoder $\text{E}_{\text{oct2octa}}$ and obtain the latent feature vector $\mathbf{u}_{\text{oct2octa}}$. Then, similar to Eq.~\ref{eq:vq}, quantized features $\mathbf{{q}}_{\text{oct2octa}}$ can also be obtained via codebook $\mathcal{Z}_{\text{oct2octa}}$. Afterward, the translated 3D OCTA image $\hat{X}_{\text{oct2octa}}$ can be obtained via decoder $\text{D}_{\text{oct2octa}}$. 
Now, we present the details of contrastive-inspired semantic alignment from 3D OCT view. 

As shown in Fig~\ref{fig:overview}, for compute efficiency, our patch operator divides $\mathbf{u}_{\text{oct2octa}}$ and $\mathbf{u}_{\text{oct}}$ into non-overlapping patch features $\mathbf{{u}_{\text{oct2octa}}}_{ \{(1,1), (1,2), \cdots, (\frac{W}{S},\frac{H}{S}) \}}$ and ${\mathbf{u}_{\text{oct}}}_{ \{(1,1), (1,2), \cdots, (\frac{W}{S},\frac{H}{S}) \}}$, where $S$ denotes the patch height and width. Moreover, we add two additional projection modules $\phi_{\text{oct2octa}}(\cdot)$ and $\phi_{\text{oct}}(\cdot)$, each of which includes a global average pooling layer and a fully connected layer. $\phi_{\text{oct2octa}}(\cdot)$ and $\phi_{\text{oct}}(\cdot)$ act as a projection layer that transforms the unquantized features from the translation model and pre-trained 3D OCT model into $\mathbf{P_{ \{(1,1), (1,2), \cdots, (\frac{W}{S},\frac{H}{S}) \}}}$ and $\mathbf{Q_{ \{(1,1), (1,2), \cdots, (\frac{W}{S},\frac{H}{S}) \}}}$, respectively. The transformed embeddings are then used for contrastive learning. 

Now, considering each anchor embedding $\mathbf{Q_{(i, j)}}$  $\in$ $\mathbf{Q_{\{(1,1), (1,2), \cdots, (\frac{W}{S},\frac{H}{S}) \}}}$, we construct the positive contrastive embeddings $\mathbf{P_{(i, j)}}$ and negative embeddings $\mathbf{P_{\{(1,1), (1,2), \cdots, (m,n),(\frac{W}{S},\frac{H}{S}) \}}}$, where $m \neq i$ or $n \neq j$. Note that, the above feature embeddings are preprocessed by $l_{2}$-normalization for numerical stability. Then, the contrastive probability distribution $\mathcal{P}_{ \text{oct2octa} \rightarrow  \text{oct}}$ between unquantized features from the translation model and pre-trained OCT reconstruction model can be formulated as:
\begin{equation}
\mathcal{P}_{ \text{oct2octa} \rightarrow  \text{oct}} = {softmax}([ \cdots, \left(\mathbf{Q_{(i, j)}}\cdot\mathbf{P_{(m, n)}} / \tau \right), \cdots ])
\label{eq:distribution} 
\end{equation}
where $\tau$ is a constant temperature and  $\mathcal{P}_{ \text{oct2octa} \rightarrow  \text{oct}} \in \mathbb{R}^{\frac{W}{S}  \cdot \frac{H}{S}}$. Notably, the unquantized features from the pre-trained OCT model are supposed to contain sufficient OCT information for its reconstruction. Thus, the translation VQVAE model can benefit from the pre-trained OCT reconstruction model by enforcing a large similarity at the same location $(i, j)$. To achieve this effect, we use the cross-entropy loss to force the positive pair to have a larger similarity upon the contrastive distribution:
\begin{equation}
\mathcal{L}_{\text{OCT}} =- \operatorname{log} \frac{\operatorname{exp}(\mathbf{Q_{(i, j)}}\cdot\mathbf{P_{(i, j)}} / \tau)}{\sum_{m=1,n=1}^{\frac{W}{S},\frac{H}{S}}\operatorname{exp}(\mathbf{Q_{(i, j)}}\cdot\mathbf{P_{(m, n)}} / \tau)}.
\label{eq:OCT} 
\end{equation}   
Meanwhile, our contrastive learning can also be performed on the quantized features from the translation model and pre-trained OCTA reconstruction model. Correspondingly, we can construct the contrastive probability distribution $\mathcal{P}_{ \text{oct2octa} \rightarrow  \text{octa}}$, and then derive $\mathcal{L}_{\text{OCTA}}$ that achieves the alignment to the pre-trained OCTA features. Compared with vanilla InfoNCE, our $\mathcal{L}_{\text{OCT}}$ and $\mathcal{L}_{\text{OCTA}}$ employs contrastive embeddings across different tasks. It can model explicit relationships between the unquantized and quantized features under the vector-quantization process, enforcing the codebook to explore more new codewords to fill the huge domain gap between OCT and OCTA modality. Fig.~\ref{fig:mo} shows that the proposed method can enhance codebook learning with an impressive high utilization rate than the vanilla VQVAE. This further verifies that, by leveraging informative unquantized and quantized features from pre-trained models, the vanilla VQVAE model can learn the codebook effectively and improve the utilization of code items significantly. 

\noindent\textbf{Theoretical guarantee.} To understander Eq.~\ref{eq:OCT} better, we reformulate mutual information in \cite{yang2023online} for image translation tasks to prove that minimizing Eq.~\ref{eq:OCT} is equal to maxmize the upper bound on the mutual information 
$I(\mathbf{P}, \mathbf{Q})$. Herein, a new proof can be derived as follows:
\begin{equation}
I(\mathbf{P}, \mathbf{Q}) \geq \log (\frac{W}{S} \cdot \frac{H}{S} -1) - \mathbb{E}_{\left (\mathbf{P}, \mathbf{Q}\right)} \mathcal{L}_{\text{OCT}}
\label{eq:info}
\end{equation}   
$\mathcal{L}_{\text{OCT}}$ maximize the information overlap degree between unquantized features from the translation and pre-trained OCT model. Meanwhile, $\mathcal{L}_{\text{OCTA}}$ maximize the information overlap degree between quantized features from the translation model and the pre-trained OCTA model. In return, the translation VQVAE model could gain extra contrastive knowledge from the pre-trained reconstruction model. 

\noindent\textbf{Vessel structure alignment:} Note that, the 2D OCTA projection map is capable of visualizing vascular structures. Given a 3D OCTA volume $X_{\text{octa}} \in \mathbb{R}^{L \times W \times D}$, it can be obtained by averaging the values of pixels along the $D$ dimension and formulated as $X^\text{proj}_{\text{octa}} \in \mathbb{R}^{L \times W}$. Importantly, Fig.~\ref{fig:vis} shows that there exist some vessel discontinuity regions in real OCTA projection maps compared with the reconstructed OCTA projection maps by the pre-trained OCTA model. The reason is that the real OCTA projection maps suffer from unstable scanning by OCTA devices. Therefore, the vanilla VQVAE may suffer from the overfitting problem by simply learning these specific patterns with only the supervision from real OCTA. To alleviate this issue, we propose vessel structure alignment from the 2D OCTA projection map view, enforcing the codebook to capture OCTA vessel continuity information. 
\begin{figure}[h]
\centering
\vspace{-4mm}		
  \begin{tabular}{cc}
\hspace{-3mm}\includegraphics[scale=0.29]{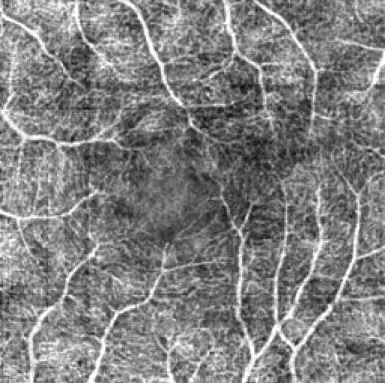}&
\includegraphics[scale=0.29]{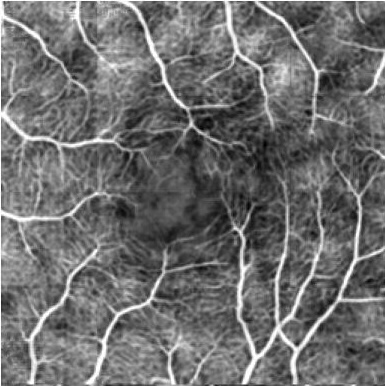}\\
     \hspace{-3mm} (a) Real OCTA PM &(b) Reconstructed OCTA PM
\end{tabular}
\vspace{-4mm}	
\caption{Visualizations of real and fake OCTA project map (PM).}
 \label{fig:vis}
\end{figure}

First, given the predicted 2D OCTA projection map $\hat{X}^\text{proj}_{\text{octa}}$ from the pre-trained OCTA  model, we divide it into the non-overlapping patch and resize each patch into one dimension vectors, formulated as $\hat{X}^\text{proj}_{ \{(1,1), (1,2), \cdots, (\frac{W}{S},\frac{H}{S}) \}}$, where $S$ denotes the patch height and width. Then, we calculate the cosine similarity ${\mathcal{C}}^{\text{octa}}_{i*\frac{H}{S} + j, m*\frac{H}{S} + n}$ between two different resized $\hat{X}^\text{proj}_{(i,j)}$ and  $\hat{X}^\text{proj}_{(m,n)}$ as follows:
\begin{equation}
\begin{aligned}
{\mathcal{C}}^{\text{octa}}_{i*\frac{H}{S} + j, m*\frac{H}{S} + n} &=\operatorname{CosSim}\left(\hat{X}^\text{proj}_{(i,j)}, \hat{X}^\text{proj}_{(m,n)}\right) \\
&=\frac{\hat{X}^\text{proj}_{(i,j)} \cdot \hat{X}^\text{proj}_{(m,n)} }{\left\| \hat{X}^\text{proj}_{(i,j)} \right\|\left\|\hat{X}^\text{proj}_{(m,n)}\right\|}, i \neq m ~\operatorname{or}~~j \neq n. 
\end{aligned}
\end{equation}   
Now, given the predicted 2D OCTA projection map $\hat{X}^\text{proj}_{\text{oct2octa}}$ from the  translation VQVAE model, the cosine similarity ${\mathcal{C}}^{\text{oct2octa}}_{i*\frac{H}{S} + j, m*\frac{H}{S} + n} $ can be obtained by the same way. Then, the patch-level semantic correlation consistency can be achieved by the following equation:
\begin{equation}
\begin{aligned}
\mathcal{L}_{\text{proj}} &= \\
&\sum_{i,j = 1,1}^{\frac{W}{S},\frac{H}{S}}\sum_{m,n=1,1}^{\frac{W}{S},\frac{H}{S}} \| {\mathcal{C}}^{\text{octa}}_{i*\frac{H}{S} + j, m*\frac{H}{S} + n}  - {\mathcal{C}}^{\text{oct2octa}}_{i*\frac{H}{S} + j, m*\frac{H}{S} + n}  \|.
\end{aligned}
\end{equation}   
With the help of alignment via $\mathcal{L}_{\text{proj}}$, the translation VQVAE model can further benefit from the pre-trained OCTA model by capturing vessel continuity information corresponding to the prior vessel structural 2D OCTA project map.

\noindent\textbf{Overall loss function.} Finally, the overall objective $\mathcal{L}_{\text{stage2}}$ is formulated as:
\begin{equation}
\begin{aligned}
\mathcal{L}_{\text{stage2}} = \mathcal{L}^{\text{oct2octa}}_{\text{VQVAE}}  (\mathcal{Z}, & \text{E}_{\text{oct2octa}},  \text{D}_{\text{oct2octa}}) \\
&+ \lambda ( \mathcal{L}_{\text{OCT}} +  \mathcal{L}_{\text{OCTA}} +  \mathcal{L}_{\text{proj}} ),
\label{eq:stage2}
\end{aligned}
\end{equation}
where $\mathcal{L}^{\text{oct2octa}}_{\text{VQVAE}}(\mathcal{Z}, \text{E}_{\text{oct2octa}}, \text{D}_{\text{oct2octa}})$ is  formulated as in Eq.~\ref{eq:vqvae}. The $\lambda$ is used to balance the above loss functions.
\section{Experiments}
\label{sec:exp}
\noindent\textbf{OCTA-3M.}  OCTA-3M is a subset of a publicly accessible dataset OCTA-500 \cite{li2020ipn}. It contains 200 pairs of OCT and OCTA volumes with 3 mm $\times$ 3 mm $\times$  2 mm field of view. The size of each volume is 304 $\times$ 304 $\times$ 640. Correspondingly, the size of the projection map is 304 $\times$  304. Following the existing work \cite{li2020ipn}, 140 volumes, 10 volumes, and 50 volumes are used as the training, validation, and test set.

\noindent\textbf{OCTA-6M.}  OCTA-6M is another subset of a publicly accessible dataset OCTA-500 \cite{li2020ipn}. It contains 300 pairs of OCT and OCTA volumes with 6 mm $\times$ 6 mm $\times$  2 mm field of view. The size of each volume is 400 $\times$ 400 $\times$ 640. Correspondingly, the size of the projection map is 400 $\times$  400. Following \cite{li2020ipn}, 
180 volumes, 20 volumes, and 100 volumes are used as the training, validation, and test set.

\noindent\textbf{OCTA2024.} We collect 846 pairs of OCT and OCTA volumes with a fixed field of view from real-world applications. The center cropped size of each volume is 256 $\times$ 256 $\times$ 256. Correspondingly, the size of the projection map is 256 $\times$  256. Following the train-val-split in the existing work \cite{li2020ipn}, 606 volumes, 40 volumes, and 200 volumes are used as the training, validation, and test set, respectively.

\subsection{Sensitivity Study}
\label{sec:sens}
\begin{figure}[h] 
\centering
\begin{tabular}{cc}
\hspace{-3mm}\includegraphics[scale=0.25]{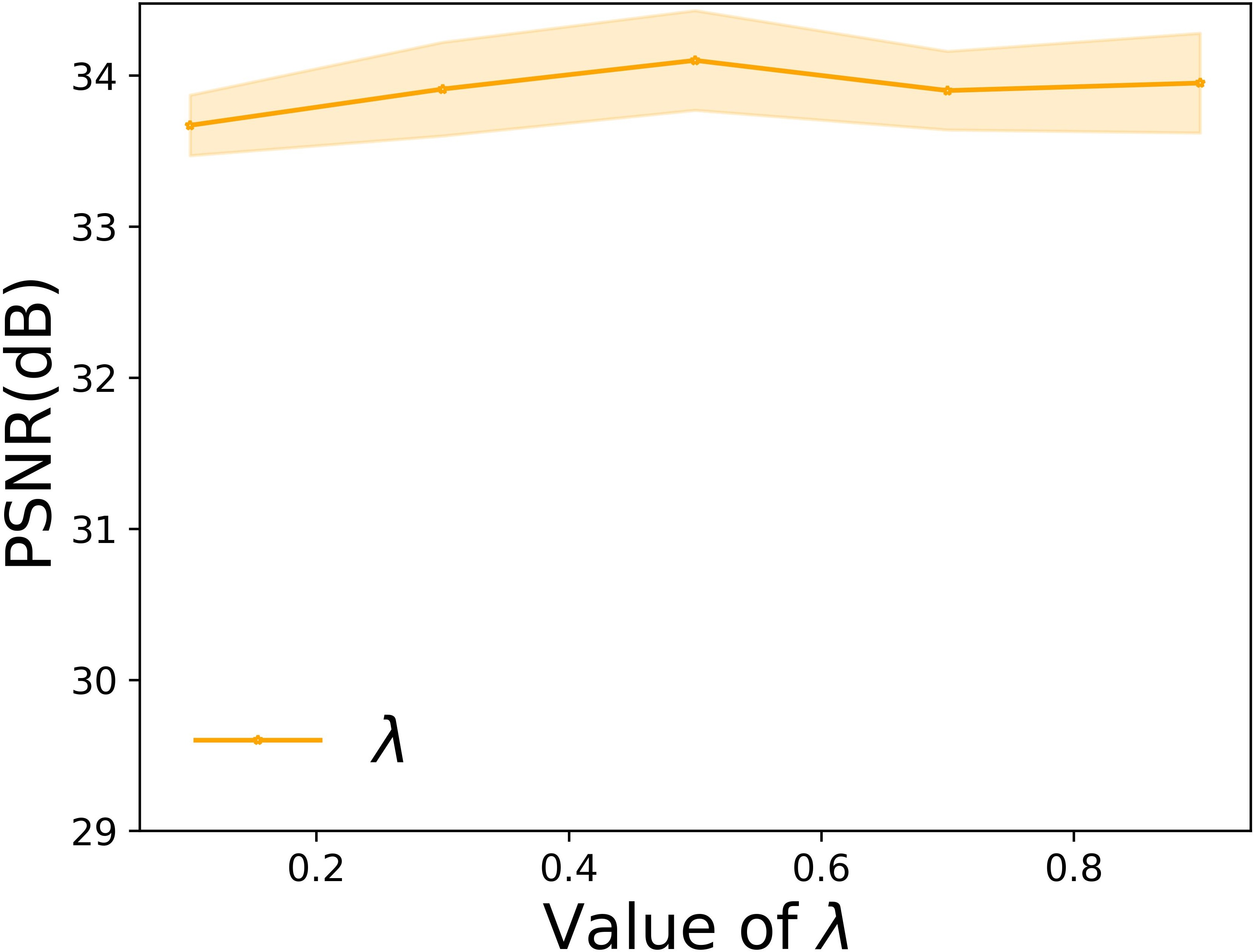}&
\hspace{-3mm}\includegraphics[scale=0.25]{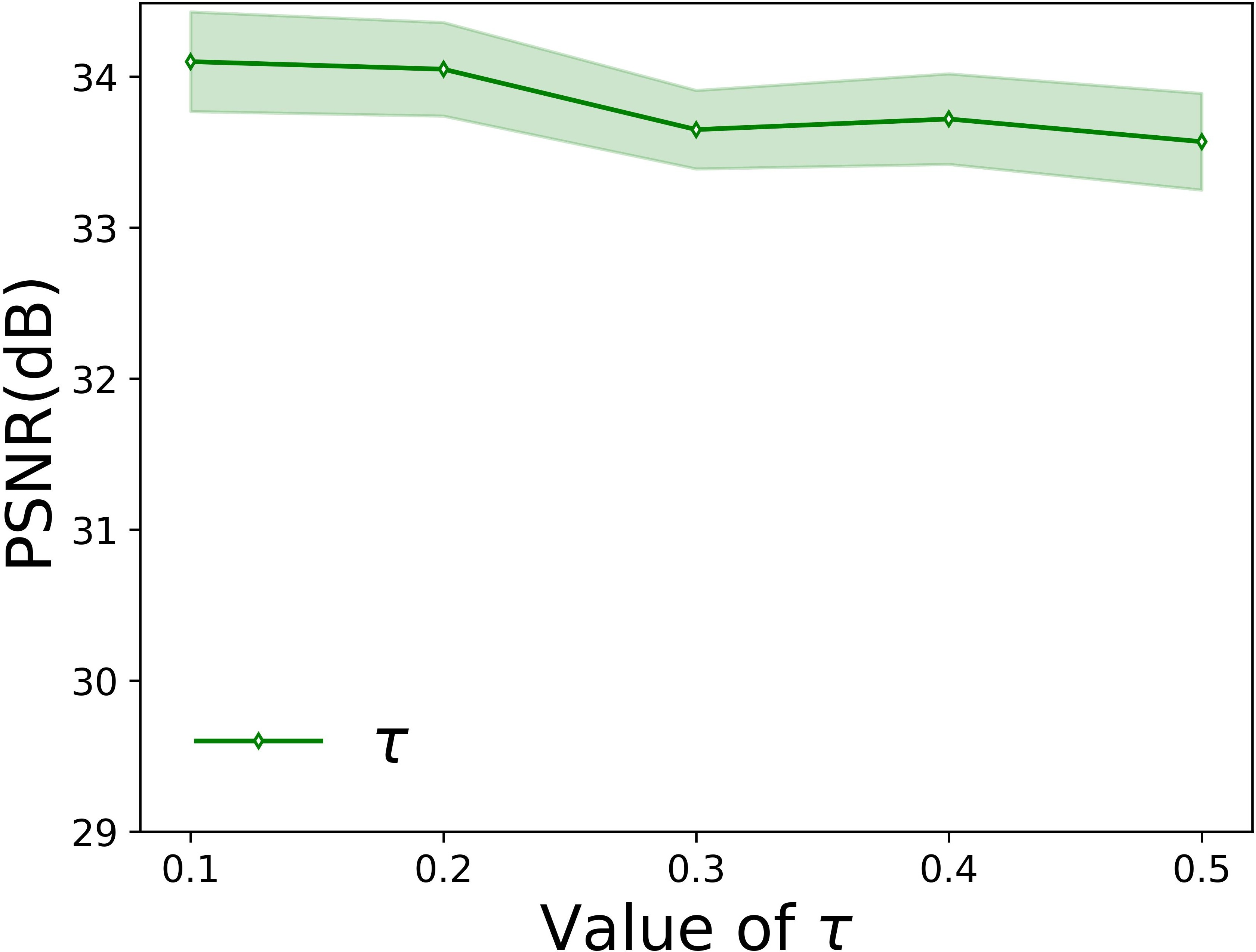}\\
\hspace{-3mm} \small{(a) Sensitivity study on $\lambda$} &\hspace{-3mm}  \small{(b) Sensitivity study on $\tau$}
\end{tabular}
\caption{Hyper-parameter sensitivity study on OCTA-3M dataset. } 
\label{fig:sens}
\end{figure}

\vspace{-1mm}
There are two hyper-parameters in this paper. $\lambda$ is used to balance the loss functions, and $\tau$ is the constant temperature in contrastive probability distribution. The hyper-parameter sensitivity study on the OCTA-3M dataset is introduced in Fig.~\ref{fig:sens}. It shows that the proposed \emph{MuTri} is not sensitive to the choice of $\lambda$ and $\tau$. Based on the PSNR metric, we set $\lambda = 0.5$ and $\tau=0.1$ in experiments.

\begin{table*}[t]
\center
\resizebox{0.97\textwidth}{!}{ 
\begin{tabular}{c|c|ccc|ccc} 
\toprule[2pt] 
{\multirow{2}{*}{ \textbf{Method}}} &\multirow{2}{*}{\textbf{Venue}}& \multicolumn{3}{c|}{ \textbf{OCTA-3M} } & \multicolumn{3}{c}{ \textbf{OCTA-6M} } \\
\cline{3-8} &   & \textbf{MAE $\downarrow$} & \textbf{PSNR(dB) $\uparrow$} & \textbf{SSIM(\%)  $\uparrow$}  & \textbf{MAE $\downarrow$} & \textbf{PSNR(dB) $\uparrow$} & \textbf{SSIM(\%)  $\uparrow$}  \\ \hline
\hline
{Pix2Pix2D \cite{isola2017image}}& \text{CVPR}& 0.0968  &  29.59 &84.59  &0.0995  &27.65  &87.15  \\
{Pix2Pix3D \cite{isola2017image}}& \text{CVPR}&0.0883  &  31.58& 86.24 &  0.0900& 30.66 &87.16  \\
 {9B18CN UNet  \cite{lee2019generating}} & - & 0.0918 & 29.34 & 86.04 & 0.1135 &27.91 & 83.69  \\
{Adjacent GAN \cite{li2020deep}} & - &  0.0906& 30.89  &87.14  &  0.1021& 28.05 &  85.03\\
{VQ-I2I \cite{chen2022eccv}} & \text{ECCV} &0.0824  & 31.72 &  87.59& 0.0897 &29.54  & 86.90 \\
{Palette \cite{saharia2022palette}} & \text{SIGGRAPH}  &0.0814 & 32.42 &88.24  &0.0881  &30.02  &  87.13 \\
{TransPro \cite{li2024vessel}} &\text{MIA}  &0.0782  &32.56  & 88.22 & 0.0854 & 30.53 & 88.35  \\
\hline
\hline
  \emph{MuTri} & -&\textbf{0.0707} & \textbf{34.10} & \textbf{89.86} & \textbf{ 0.0741} & \textbf{ 33.08 }  & \textbf{90.04}  \\
\toprule[1pt] 
\end{tabular}
} 
\caption{Comparisons with the state-of-the-art  OCT to OCTA image translation methods on both OCTA-3M and OCTA-6M datasets.}
\label{tab:OCTA}
\end{table*}

\begin{table*}[t]
\center
\resizebox{0.97\textwidth}{!}{ 
\begin{tabular}{c|c|ccc|ccc} 
\toprule[2pt] 
{\multirow{2}{*}{ \textbf{Method}}} &\multirow{2}{*}{\textbf{Venue}}& \multicolumn{3}{c|}{ \textbf{OCTA2024 (OCTA Volume)} } & \multicolumn{3}{c}{ \textbf{OCTA2024 (OCTA Projection Map)} } \\
\cline{3-8} &   & \textbf{MAE $\downarrow$} & \textbf{PSNR(dB) $\uparrow$} & \textbf{SSIM(\%)  $\uparrow$}  & \textbf{MAE $\downarrow$} & \textbf{PSNR(dB) $\uparrow$} & \textbf{SSIM(\%)  $\uparrow$}  \\ 
\hline
\hline
{Pix2Pix2D \cite{isola2017image}}& \text{CVPR}&   0.1301 &   40.65& 69.78 &  .01471& 35.17 & 86.39  \\
{Pix2Pix3D \cite{isola2017image}}& \text{CVPR}& 0.1275   &  41.87 & 71.92 &.01324  & 36.88 &86.74   \\
 {9B18CN UNet  \cite{lee2019generating}} & - & 0.1402   & 40.53  &70.76  & .01586 & 34.79 &84.37   \\
{Adjacent GAN \cite{li2020deep}} & - & 0.1352   & 41.97  & 71.43 & .01459 & 36.18 & 85.97  \\ 
{VQ-I2I \cite{chen2022eccv}} & \text{ECCV} & 0.1187   &41.25   &71.31  &  .01132& 37.29 & 87.24  \\
{Palette \cite{saharia2022palette}} & \text{SIGGRAPH}  &  0.1254  &41.40   & 71.07 &.01241  &  37.54&87.33   \\
{TransPro \cite{li2024vessel}} &\text{MIA}  &   0.1098 & 42.69  & 72.52 &.01056 & 37.85 &  87.61 \\
\hline
\hline
 \emph{MuTri} & -&\textbf{0.0828 } & \textbf{43.38} & \textbf{79.66} & \textbf{.00870 } & \textbf{  39.65}  & \textbf{90.31}  \\
\toprule[1pt] 
\end{tabular}
} 
\caption{Comparisons with the state-of-the-art OCT to OCTA  translation methods on our proposed OCT2024 dataset and its project map.}
\label{tab:OCTA2024}
\end{table*}

\subsection{Implementation Details} 
Our \emph{MuTri} is implemented based on the PyTorch \cite{paszke2019pytorch}. Followed by \cite{roy2023mednext}, the encoder/decoder of \emph{MuTri} consists of four blocks, where each block contains two ResBlocks \cite{he2016deep} and a downsampling/upsampling layer. 
Each downsampling layer is followed by a codebook. For both two training stages, we use an Adam optimizer \cite{diederik2014adam} with a batch size of 1, a learning rate of 3.0 $\times$ 10$^{-4}$, and the training epoch of 2000.  According to our sensitivity study in Sec.~\ref{sec:sens}, $\tau$ and $\lambda$ are set as 0.1 and 0.5.

\noindent\textbf{Evaluation metrics}  We evaluate the translated OCTA images from the following two aspects: OCTA volume and OCTA projection map. To evaluate the quality of OCTA volume, we use standard image quality assessment metrics, i.e., Mean Absolute Error (MAE), Peak Signal-to-Noise Ratio (PSNR), and Structural SIMilarity (SSIM) \cite{wangzhou2004image} and average the results over each slide of the whole volume \cite{li2024vessel}. 

\subsection{Quantitative and  Diagnosis Analysis}

\noindent\textbf{OCTA-3M.} The re-implemented results for the OCTA-3M dataset are displayed in Table~\ref{tab:OCTA}. From the results, we observe that pix2pix3D generally outperformed pix2pix2D. The reason is that the inter-slice connections in 3D volume can provide more global information compared to a single B-scan image. Although Palette \cite{saharia2022palette} shows the advantage of utilizing diffusion models for image translation tasks, it involves longer inference times due to its sophisticated denoising steps. The latest approaches, TransPro \cite{li2024vessel} incorporate vessel and contextual guidance from the OCTA project map view, leading to an improvements in the OCTA-3M dataset compared to other previous methods. In contrast to these methods, our \emph{MuTri} leverages multi-view guidance from 3D OCT, 3D OCTA, and 2D OCTA projection map, and learn the OCT to OCTA mapping in discrete and finite space. Notably, in Table 1, our method demonstrates significant improvement over
the TransPro by 0.75\% in MAE, 1.54 dB in PSNR, and 1.64\% in SSIM.

\noindent\textbf{Diagnosis analysis.} Herein, to further verify the potential practical applications of our proposed method, we cooperate with ophthalmologists to analyze the abnormal decreased capillary density diseased patterns in both real and translated OCTA projection maps. We compare the real OCTA images, the translation results from TransPro, and our \emph{MuTri}. Our findings reveal that major diseased patterns are preserved in the translated OCTA projection maps, albeit with some ambiguity in the details of fine vessel areas. Moreover, compared to the TransPro, our proposed method generates OCTA projection maps that are more similar to the real ones in terms of diseased levels, as shown in Fig.~\ref{fig:vis}. Note that, the content has been generated by our proposed generative model. We do not assume any responsibility or liability for the use or interpretation of this content. 
\begin{figure}[h] 
\centering
\includegraphics[scale=0.43]{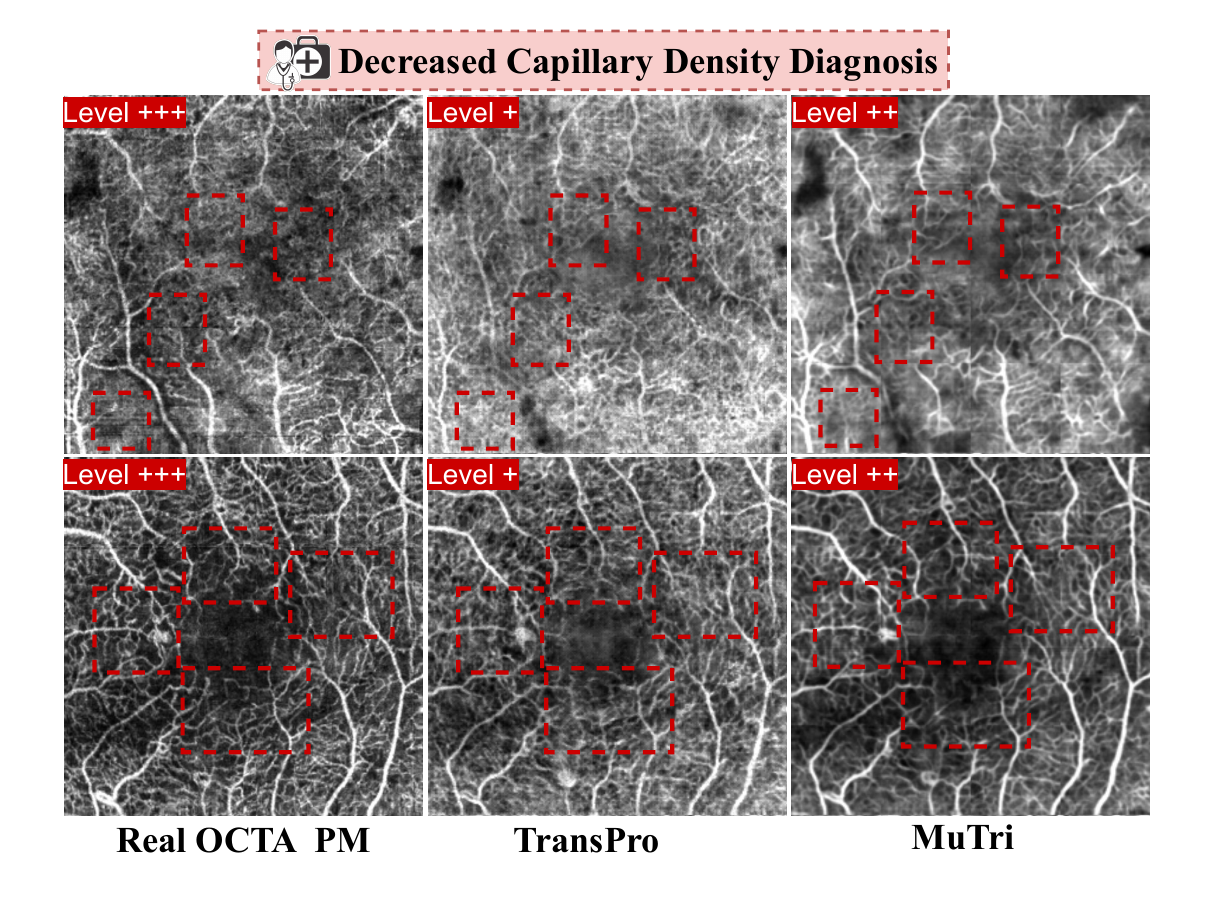}
\caption{The decreased capillary density diseased patterns of translated OCTA projection maps (PM) annotated by an experienced ophthalmologist. The diseased pattern is presented alongside the real OCTA image, the translation results from TransPro, and our \emph{MuTri}. Levels of disease are annotated from ‘‘+’’ to ‘‘+++’’, where the number of ‘‘+’’ denotes the degree of severity.} 
\label{fig:vis}
\end{figure}

\noindent\textbf{OCTA-6M.} As we can see from Table~\ref{tab:OCTA}, the proposed method achieves the best performance on all three metrics, i.e., 0.0741 in MAE, { 33.08 } dB in PSNR, and {90.04} in SSIM metrics, even facing the lower resolution of images with the larger field of view in the OCTA-6M dataset. This further verifies the pre-trained model can provide high-quality semantic priors. Meanwhile, the single-view guidance of TransPro \cite{li2024vessel} contributes the { 0.0854} in MAE, { 30.53 } dB in PSNR, and {88.35} in SSIM metrics. The reason behind this is that TransPro introduces vascular segmentation as an auxiliary task, while the segmentation performance can not be guaranteed all the time, and its pre-trained vascular segmentation model involves huge pixel-level labeling costs.

\noindent\textbf{OCTA2024.}~The re-implemented results for the OCTA2024 dataset are displayed in Table~\ref{tab:OCTA2024}. It shows that the proposed method again achieves the best performance on all six metrics, i.e., 0.0828 in MAE, {43.38} dB in PSNR, {79.66} in SSIM metrics on OCTA volume. And, it can be observed that our \emph{MuTri} outperforms the existing work by a large margin. The reason can be attributed to two-folds: (i) Our large-scale OCTA2024 dataset can promote the pre-trained models to produce high-quality semantic prior, which has been overlooked by the previous methods. Although TransPro \cite{li2024vessel} also leverage the pre-trained segmentation model, its generalization capability to our large-scale dataset can not be guaranteed. (ii) Our method 
 advances contrastive learning with multi-view guidance for the OCT to OCTA translation task. \emph{MuTri} performs contrastive learning across self-reconstruction and image translation tasks to maximize the mutual information with the pre-trained OCT and OCTA reconstruction model. In this way, we can easily learn the mapping with multi-view guidance.

\subsection{Ablation Studies and Scalability Discussion}
We further conduct ablation studies to evaluate our two proposed alignments: contrastive-inspired semantic alignment (CSA) and vessel structure alignment (VSA), which are verified on OCTA-6M datasets. We study the importance of the above two proposed alignments, as shown in Table~\ref{tab:abla}. It can be seen that our proposed CSA has significantly improved performance. In addition, with the proposed VSA, the performance can be further improved. Thus, we can see that it is crucial to learn vessel structure information from the 2D OCTA project map view. We believe our solution can inspire other 3D medical translation tasks, i.e., CT-to-MRI, as our proposed method reveals that the translation model can benefit from the pre-trained model via our CSA for the alignment of unquantized/quantized features from a mutual information perspective. 

\begin{table}[h]
\center
  \resizebox{0.47\textwidth}{!}{ 
\begin{tabular}{c|cc|ccc}
\toprule[2pt]
 \multirow{2}{*}{Method}& \multicolumn{2}{c|}{ \textbf{Alignment}}& \multirow{2}{*}{\textbf{MAE}} & \multirow{2}{*}{\textbf{PSNR(dB)}} & \multirow{2}{*}{\textbf{SSIM(\%)}}  \\
 \cline{2-3}  & \text{CSA} & \text{VSA}  &&& \\  
\toprule[1pt]
 \multirow{3}{*}{\emph{MuTri}}  &-  &  \checkmark & 0.0765 & 32.31   & 89.53  \\ 
  & \checkmark  &- &  0.0748 & 32.87 &89.19  \\ 
    & \checkmark  &  \checkmark & \textbf{ 0.0741} & \textbf{ 33.08 }  & \textbf{90.04}   \\ 
\toprule[2pt]  
\end{tabular}
}
\caption{Ablation studies: ``CSA'' denotes contrastive-inspired semantic alignment and ``VSA'' denotes vessel structure alignment.}
\label{tab:abla}
\end{table}

\section{Conclusion}
In this paper, we propose Multi-view Tri-alignment, namely \emph{MuTri} for OCT to OCTA 3D image translation in discrete and finite space. Motivated by the fact that pre-trained OCT and OCTA reconstruction models can produce high-quality semantic prior at triple views: 3D OCT, 3D OCTA, and 2D OCTA project maps,  we propose multi-view alignments to promote the optimization of codebook learning and enforce it to capture the detailed vessel structure information. Specifically, we propose contrastive-inspired semantic alignment that maximizes the mutual information with the pre-trained OCT and OCTA reconstruction models from 3D OCT and 3D OCTA views. Meanwhile, vessel structure alignment is proposed to achieve vessel structure consistency from the 2D OCTA projection map view. This way, the translation VQVAE model can benefit from the pre-trained reconstruction models and achieve superior performance on two publicly available datasets and our proposed first large-scale datasets. 

\section{Acknowledgements}
This work is supported by research grants from the Foshan HKUST Projects (Grants FSUST21-HKUST10E and FSUST21-HKUST11E) and the Guangdong Provincial Science and Technology Fund (Project 2023A0505030004).

\clearpage
{
    \small   
    \bibliographystyle{ieeenat_fullname}
    \bibliography{OCT2OCTA}
}

\end{document}